\pdfoutput=1
\documentclass{article}

\usepackage{microtype}
\usepackage{graphicx}
\usepackage{subfigure}
\usepackage{booktabs} 

\usepackage{hyperref}

\newcommand{\dds}{DDS}

\usepackage[linesnumbered,ruled]{algorithm2e}
\usepackage{adjustbox}
\usepackage{mathtools}
\usepackage{multirow}
\usepackage{wrapfig}

\usepackage[accepted]{icml2019}

\icmltitlerunning{Optimizing Data Usage via Differentiable Rewards}
\usepackage{times}
\usepackage{float}
\usepackage{graphicx}
\usepackage{amsmath}
\usepackage{subfigure}
\usepackage{microtype}
\usepackage{url}
\usepackage{amsfonts}
\usepackage{amssymb}
\usepackage{amsthm}
\usepackage{mathrsfs}
\usepackage{tikz}
\usepackage{graphicx}
\usepackage{verbatim}
\usepackage{xcolor}
\usepackage{booktabs}
\usepackage{multirow}
\usepackage{todonotes}
\usepackage{footnote}
\usepackage{wrapfig}
\usepackage{blindtext}
\usepackage[font=small,labelfont=bf]{caption}
\usepackage{mwe}
\usepackage{algorithmic}

\makesavenoteenv{tabular}
\usepackage[toc,page]{appendix}
\usepackage{comment}

\newcommand{\eg}{\textit{e.g.}}
\newcommand{\ie}{\textit{i.e.}}

\newcommand{\ABS}[1]{\left| #1 \right|}

\DeclareMathOperator*{\argmin}{argmin}
\begin{document}

\twocolumn[
\icmltitle{Optimizing Data Usage via Differentiable Rewards}



\icmlsetsymbol{equal}{*}

\begin{icmlauthorlist}
\icmlauthor{Xinyi Wang}{equal,cmu}
\icmlauthor{Hieu Pham}{equal,cmu,goo}
\icmlauthor{Paul Michel}{cmu}
\icmlauthor{Antonios Anastasopoulos}{cmu}
\icmlauthor{Jaime Carbonell}{cmu}
\icmlauthor{Graham Neubig}{cmu}
\end{icmlauthorlist}

\icmlaffiliation{cmu}{Language Technology Institute, Carnegie Mellon University, Pittsburgh, PA 15213, USA}
\icmlaffiliation{goo}{Google Research, Brain Team, Mountain View, CA 94043, USA}

\icmlcorrespondingauthor{Xinyi Wang, Hieu Pham}{\{xinyiw1,hyhieu\}@cs.cmu.edu}

\icmlkeywords{Machine Learning, ICML}

\vskip 0.3in
]



\printAffiliationsAndNotice{\icmlEqualContribution} 

\begin{abstract}
To acquire a new skill, humans learn better and faster if a tutor informs them of how much attention they should pay to particular content or practice problems  based on their current knowledge level. Similarly, a machine learning model could potentially be trained better if data is presented in a way that adapts to its current learning state.
In this paper, we examine the problem of training an adaptive scorer that weights data instances to maximally benefit learning.
Training such as scorer efficiently is a challenging problem; in order to precisely quantify the effect of a data instance on the final model, a naive approach would require completing the entire training process and observing final performance.
We propose an efficient alternative -- Differentiable Data Selection~(\dds) -- that formulates a scorer as a learnable function of the training data that can be efficiently updated along with the main model being trained. Specifically, \dds~updates the scorer with an intuitive reward signal: it should up-weigh the data that has a similar gradient with a development set upon which we would finally like to perform well. Without significant computing overhead, \dds~delivers consistent improvements over several strong baselines on two very different tasks of machine translation and image classification.%

\end{abstract}
\section{\label{sec:intro} Introduction}



While deep learning models are remarkably good at fitting large data sets, their performance is also highly sensitive to the structure and domain of their training data. Training on out-of-domain data can lead to worse model performance, while using more relevant data can assist transfer learning.
Previous work has attempted to create strategies to handle this sensitivity by selecting subsets of the data to train the model on \citep{jiang-zhai-2007-instance,wang-etal-2017-instance,axelrod2011domain,moore2010intelligent}, providing different weights for each example \citep{importance_weight,learn_reweight}, or changing the presentation order of data \citep{cl_bengio,rl_nmt}.

However, there are several challenges with existing work on better strategies for data usage. Most data filtering critera or training curriculum rely on domain-specific knowledge and hand-designed heuristics, which can be sub-optimal. To avoid hand-designed heuristics, several works propose to optimize a parameterized neural network to learn the data usage schedule, but most of them are tailored to specific use cases, such as handling noisy data for classification~\citep{mentornet}, learning a curriculum learning strategy for specific tasks~\citep{rl_nmt,baysian_curriculum}, and actively selecting data for annotation~\citep{learn_active_learn,reinforce_cotrain}.
\citet{learn_to_teach}~proposes a more general teacher-student framework that first trains a ``teacher network'' to select data that directly optimizes development set accuracy over multiple training runs.
However, because running multiple runs of training simply to learn this teacher network entails an $n$-fold increase in training time for $n$ runs, this is infeasible in many practical settings.
In addition, in preliminary experiments we also found the single reward signal provided by dev set accuracy at the end of training to be noisy, to the extent that we were not able to achieve results competitive with simpler heuristic training methods.

\begin{figure}
    \centering
    \includegraphics[width=0.4\textwidth]{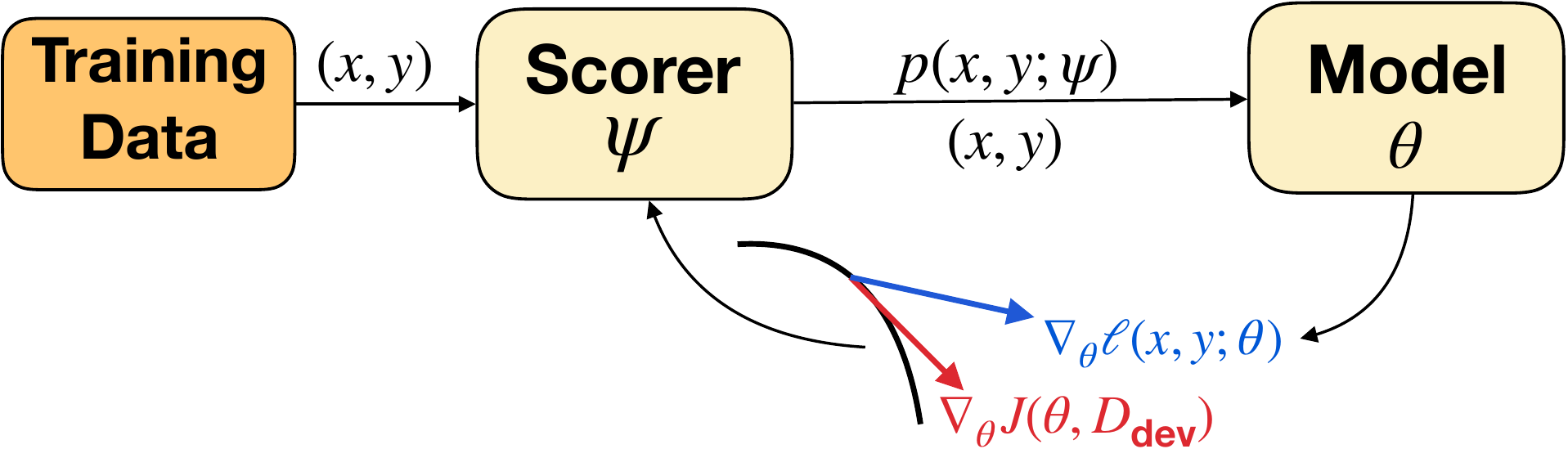}
    \caption{The general workflow of \dds.}
    \label{fig:method}
\end{figure}

In this paper, we propose an alternative: a general Reinforcement Learning~(RL) framework for optimizing training data usage by training a \emph{scorer network} that minimizes the model loss on the development set.
We formulate the scorer network as a function of the current training examples, and update the scorer along with the main model being trained. Thus, our method requires no heuristics and is generalizable to various tasks. To make the scorer adaptive, we train it using frequent and efficient updates towards a reward function inspired by recent work on learning from auxiliary tasks~\citep{cos_sim,meta_aux_learn}, which uses the similarity between two gradients as a measure of task relevance.
We propose to use the gradient alignment between the training examples and the dev set as a reward signal for \emph{a parametric scorer network}, as illustrated in \autoref{fig:method}. We then formulate our framework as a bi-level optimization problem similar to those found in prior works such as meta-learning for few-shot learning~\citep{finn2017model}, noisy data filtering~\citep{learn_reweight}, and neural architecture search~\citep{darts}, and demonstrate that our proposed update rules follow a direct differentiation of the scorer parameters to optimize the model loss on the dev set.
Thus we refer to our framework as ``Differentiable Data Selection''~(\dds).

We demonstrate two concrete instantiations of the \dds~framework, one for a more general case of image classification, and the other for a more specific case of neural machine translation~(NMT). For image classification, we test on both CIFAR-10 and ImageNet. For NMT, we focus on a multilingual setting, where we optimize data usage from a multilingual corpus to improve the performance on a particular language. 
For these two very different and realistic tasks, we find the \dds~framework brings significant improvements over diverse baselines for all settings.
\section{\label{sec:method} Differentiable Data Selection}

\subsection{\label{sec:dds_motivation}Risk, Training, and Development Sets}

Commonly in machine learning, we seek to find the parameters $\theta^*$ that minimize the \emph{risk} $J(\theta,P)$, the expected value of a loss function $\ell(x, y; \theta)$, where $\langle x, y \rangle$ are pairs of inputs and associated labels sampled from a particular distribution $P(X, Y)$:
\begin{equation}
  \label{eqn:generic_optim}
  \begin{aligned}
   & \theta^* = \argmin_\theta J(\theta, P)
    ~~~\text{where}~~~ \\
   & J(\theta, P) = \mathbb{E}_{x, y \sim P(X, Y)} [\ell(x, y; \theta)]
  \end{aligned}
\end{equation}

Ideally, we would like the risk $J(\cdot)$ to be minimized over the data distribution that our system sees at test time, ie.~$P_{\text{test}}(X,Y)$.
Unfortunately, this distribution is unknown at training time, so instead we collect a training set $\mathcal{D}_\text{train} = \{(x_i, y_i): i = 1, ..., N_\text{train}\}$ with distribution $P_\text{train}(X, Y) = \text{Uniform}(\mathcal{D}_\text{train})$, and minimize the \emph{empirical risk} by taking $\langle x, y \rangle \sim P_\text{train}(X, Y)$.
Since we need a sufficiently large training set $\mathcal{D}_\text{train}$ to train a good model, it is hard to ensure that $P_\text{train}(X, Y) \approx P_{\text{test}}(X, Y)$. In fact, we often accept that training data comes from a different distribution than test data.
The discrepancy between $P_\text{train}(X, Y)$ and $P_\text{test}(X, Y)$ manifests itself in the form of problems such as overfitting~\citep{overfit_random_examples,dropout}, covariate shift~\citep{shimodaira2000improving}, and label shift~\citep{lipton2018detecting}.

However, unlike the large training set, we can often collect a relatively small development set $\mathcal{D}_\text{dev}= \{(x_i, y_i): i = 1, ..., N_\text{dev}\}$ with a distribution $P_{\text{dev}}(X, Y)$ that is much closer to $P_{\text{test}}(X, Y)$~(Some examples can be found in \autoref{sec:experiment}).
Since $\mathcal{D}_\text{dev}$ is a better approximation of our test-time scenario,
we can use $\mathcal{D}_\text{dev}$ to get reliable feedback to learn to better utilize our training data from $\mathcal{D}_\text{train}$. In particular, we propose to train a \emph{scorer} network, parameterized by $\psi$, that adjusts the weights of examples in $\mathcal{D}_\text{train}$ to minimize $J(\theta, \mathcal{D}_\text{dev})$ .


\subsection{\label{sec:efficient_reward} Learning to Optimize Data Usage}
We propose to optimize the scorer's parameters $\psi$ in an RL setting.
Our \emph{environment} is the model state $\theta$ and an example $\langle x, y \rangle$. Our RL \emph{agent} is the scorer network $\psi$, which optimizes the data usage for the current model state. The agent's~\emph{reward} on picking an example approximates the dev set performance of the resulting model after the model is updated on this example.

Our scorer network is parameterized as a differentiable function that only takes as inputs the features of the example $\langle x, y \rangle$. Intuitively, it represents a distribution over the training data where more important data has a higher probability of being used, denoted $P(X, Y; \psi)$. Unlike prior methods which generally require complicated featurization of both the model state and the data as input to the RL agent~\citep{learn_to_teach,mentornet,learn_active_learn}, our formulation is much simpler and  generalizable to different tasks. Since our scorer network does not consider the model parameters $\theta_t$ as input, we update it iteratively with the model so that at training step $t$, $P(X, Y; \psi_t)$ provides an up-to-date data scoring feedback for a given $\theta_t$. 

Although the above formulation is simpler and more general, it requires much more frequent updates to the scorer parameter $\psi$. Existing RL frameworks simply use the change in dev set risk as the regular reward signal, which makes the update expensive and unstable~\citep{learn_to_teach,rl_nmt}. Therefore, we propose a novel reward function as an approximation to $\Delta J_{\text{dev}}(x, y)$ to quantify the effect of the training example $\langle x, y \rangle$. Inspired by \citet{cos_sim}, which uses gradient similarity between two tasks to measure the effect of adaptating between them, we use the agreement between the model gradient on data $\langle x, y \rangle$ and the gradient on the dev set to approximate the effect of $\langle x, y \rangle$ on dev set performance. This reward implies that we prefer data that moves $\theta$ in the direction that minimizes the dev set risk: 
\begin{equation}
    \label{eqn:reward_fn}
\begin{aligned}
     R(x, y) & = \Delta J_{\text{dev}}(x, y) \\
    & \approx \nabla_\theta \ell(x, y; \theta_{t-1})^\top \cdot \nabla_\theta J(\theta_t, \mathcal{D}_\text{dev}) 
\end{aligned}
\end{equation}

According to the REINFORCE algorithm~\citep{reinforce}, the update rule for $\psi$ is thus
\begin{equation}
\label{eqn:psi_update}
\begin{aligned}
    & \psi_{t+1} \leftarrow  \psi_t + \\
    &  \underbrace{\nabla_\theta \ell(x, y; \theta_{t-1}) \cdot \nabla_\theta J(\theta_t, \mathcal{D}_\text{dev})}_{\mathclap{R(x, y)}} \nabla_\psi \text{log}(P(X, Y;\psi))
\end{aligned}
\end{equation}
The update rule for the model is simply
\begin{align}
    \label{eqn:theta_update}
    \theta_t \leftarrow \theta_{t-1} - \nabla_\theta J(\theta_{t-1}, P(X, Y;\psi))
\end{align}
For simplicity of notation, we omit the learning rate term. The full derivation can be found in \autoref{app:grad_of_optimizers}. By alternating between \autoref{eqn:theta_update} and \autoref{eqn:psi_update}, we can iteratively update $\theta$ using the guidance from the scorer network, and update $\psi$ to optimize the scorer using feedback from the model.  

Our formulation of scorer network as $P(X, Y; \psi)$ has several advantages. First, it provides the flexibility that we can either (1) sample a training instance with probability proportional to its score, (2) or equivalently scale the update from the training instance based on its score. In later sections, we provide an algorithm under the \dds~framework for multilingual NMT~(see \autoref{sec:nmt_method}), where the former is more efficient, and another more general algorithm for image classification~(see \autoref{sec:image_method}), where the latter choice is natural. Second, it allows easy integration of prior knowledge of the data, which is shown to be effective in \autoref{sec:experiment}. 

\subsection{\label{sec:diff_data_selection}Deriving Rewards through Direct Differentiation}
In this section, we show that the update for the scorer network in \autoref{eqn:psi_update} can be approximately derived as the solution of a bi-level optimization problem~\citep{bilevel_optim}, which has been applied to many different lines of research in the field of meta-learning~\citep{hyper_grad,darts,learn_reweight}. 

Under our framework, the scorer samples the data according to $\langle x, y \rangle \sim P(X, Y; \psi)$, and $\psi$ will be chosen so that $\theta^*$ that minimizes $J(\theta, P(X, Y;\psi))$ will approximately minimize $J(\theta, P_\text{dev}(X,Y))$: 
\begin{equation}
  \label{eqn:psi_theta_argmin}
  \begin{aligned}
   & \psi^* = \argmin_\psi
  J(\theta^*(\psi), \mathcal{D}_\text{dev}) 
    ~\text{where}~ \\
   & \theta^*(\psi) = \argmin_\theta \mathbb{E}_{x, y \sim P(X, Y; \psi)} \left[ \ell(x, y; \theta) \right]
  \end{aligned}
\end{equation}

The connection between $\psi$ and $\theta$ in \autoref{eqn:psi_theta_argmin} shows that $J(\theta_t, \mathcal{D}_\text{dev})$ is differentiable with respect to $\psi$. Now we can approximately compute the gradient $\nabla_\psi J(\theta_t, \mathcal{D}_\text{dev})$ as follows:

\begin{equation}
  \label{eqn:two_step_update}
  \begin{aligned}
   & \nabla_\psi J(\theta_t, \mathcal{D}_\text{dev})\\
   & \quad\quad\quad\quad \text{(apply chain rule:)} \\
    &= \nabla_{\theta_t} J(\theta_t, \mathcal{D}_\text{dev})^\top \cdot \nabla_\psi \theta_t(\psi) \\
    & \quad\quad\quad\quad  \text{(substitute $\theta_t$ from \autoref{eqn:theta_update}:)} \\
      &= \nabla_{\theta_t} J(\theta_t, \mathcal{D}_\text{dev})^\top \cdot \nabla_\psi \left( \theta_{t-1} - \nabla_\theta J(\theta_{t-1}, \psi) \right)  \\
      & \quad\quad\quad\quad  \text{(assume $\nabla_\psi \theta_{t-1} \approx 0$:)} \\
      &\approx -\nabla_{\theta_t} J(\theta_t, \mathcal{D}_\text{dev})^\top \cdot \nabla_\psi  \left( \nabla_\theta J(\theta_{t-1}, \psi) \right) \\
      &= -\nabla_\psi \mathbb{E}_{x, y \sim P(X, Y; \psi)} \left[\nabla_\theta J(\theta_t, \mathcal{D}_\text{dev})^\top \cdot \nabla_\theta \ell(x, y; \theta_{t-1} )\right] \\
    &= -\mathbb{E}_{x, y \sim P(X, Y; \psi)} \Big[\left( \nabla_\theta J(\theta_t, \mathcal{D}_\text{dev})^\top \cdot \nabla_\theta \ell(x, y; \theta_{t-1} ) \right) \\
    & \quad\quad\quad\quad\quad\quad\quad\quad\quad\quad\quad\quad\quad\quad\quad \cdot \nabla_\psi \log{P(x, y; \psi)} \Big]
  \end{aligned}
\end{equation}
Here, we make a Markov assumption that $\nabla_\psi \theta_{t-1} \approx 0$, assuming that at step $t$, given $\theta_{t-1}$ we do not care about how the values of $\psi$ from previous steps led to $\theta_{t-1}$. Intuitively, this assumption indicates in the previous step $\psi_{t-1}$ is already updated regards to $\theta_{t-1}$, so the effect of $\psi$ on $\theta_{t-1}$ is likely to be minimal. This assumption can simplify and speed up computation. Moreover, this allows us to have a natural interpretation of the update rule for the data scorer: it should up-weight the training data that have similar gradient direction with the dev data\footnote{Our use of the Markov assumption is based on its use and empirical success in previous work on bi-level optimization, such as Hyper Gradient Descent (Baydin et al. 2017) and many others. Of course, this is a simplifying assumption, but we believe that our empirical results show that the proposed method is useful nonetheless.}. \autoref{eqn:two_step_update} leads to a rule to update $\psi$ using gradient descent, which is exactly the same as the RL update rule in \autoref{eqn:psi_update}.

\subsection{Additional Derivation Details and Clarifications}
Note that our derivation above does not take into the account that we might use different optimization algorithms, such as SGD or Adam~\citep{adam}, to update $\theta$. We provide detailed derivations for several popular optimization algorithms in \autoref{app:grad_of_optimizers}.     

One potential concern with our approach is that because we optimize $\psi_t$ directly on the dev set using $J(\theta_t, \mathcal{D}_\text{dev})$, we may risk indirectly overfitting model parameters $\theta_t$ by selecting a small subset of data that is overly specialized.
However we do not observe this problem in practice, and posit that this because (1) the influence of $\psi_t$ on the final model parameters $\theta_t$ is quite indirect, and acts as a ``bottleneck'' which has similarly proven useful for preventing overfitting in neural models \cite{grezl2007probabilistic}, and (2) because the actual implementations of \dds~(which we further discuss in \autoref{sec:formualtion}) only samples a subset of data from $\mathcal{D}_\text{train}$ at each optimization step, further limiting expressivity.

\section{\label{sec:formualtion}Concrete Instantiations of \dds}

We now turn to discuss two concrete instantiations of \dds~that we use in our experiments: a more generic example of classification, which should be applicable to a wide variety of tasks, and a specialized application to the task of multilingual NMT, which should serve as an example of how \dds~can be adapted to the needs of specific applications.


\subsection{\label{sec:image_method}Formulation for Classification}

\autoref{alg:image_classification_dds} presents the pseudo code for the training process on classification tasks, using the notation introduced in \autoref{sec:method}. 

\begin{algorithm}
\SetAlgoLined
\DontPrintSemicolon
\SetKwInOut{Input}{Input}
\SetKwInOut{Output}{Output}
\SetCommentSty{itshape}
\SetKwComment{Comment}{$\triangleright$\ }{}
\Input{$\mathcal{D}_\text{train}$, $\mathcal{D}_\text{dev}$}
\Output{Optimal parameters $\theta^*$}
 Initializer $\theta_0$ and $\psi_0$
 
 \For{$t = 1$~\textbf{\emph{to}}~$\text{num\_train\_steps}$}{
    Sample $B$ training data points $x_i, y_i \sim \text{Uniform}(\mathcal{D}_\text{train})$
    
    Sample $B$ validation data points $x'_i, y'_i \sim \text{Uniform}(\mathcal{D}_\text{dev})$
  
    \Comment{Optimize $\theta$}
    $g_\theta \leftarrow 
    \sum_{i=1}^{B} p(x_i, y_i; \psi_{t-1}) \nabla_\theta \ell(x_i, y_i; \theta_{t-1})$
    
    Update $\theta_t \leftarrow \text{GradientUpdate}\Big(\theta_{t-1},
    g_\theta \Big)$
    \label{alg:grad_update_model}
    
    \Comment{Evaluate $\theta_t$ on $\mathcal{D}_\text{dev}$}
    Let $d_\theta \leftarrow \frac{1}{B} \sum_{j=1}^{B} \nabla_\theta \ell(x'_j, y'_j; \theta_t)$
  
    \Comment{Optimize $\psi$}
    $r_i \leftarrow d_\theta^\top \cdot \nabla_\theta \ell(x_i, y_i; \theta_{t-1})$
    
    Let $d_\psi \leftarrow \frac{1}{B} \sum_{i=1}^{B} \left[r_i \cdot \nabla_\psi \log{p(x_i, y_i; \psi)} \right]$
    \label{alg:require_per_example_grad}
    
    Update $\psi_t \leftarrow \text{GradientUpdate}(\psi_{t-1}, d_\psi)$
    \label{alg:grad_update_p}
  }
  \caption{\label{alg:image_classification_dds}Training a classification model with \dds.}
\end{algorithm}

The main  classification model is parameterized by $\theta$. The scorer $p(X, Y; \psi)$ uses an architecture identical to the main model, but with independent weights, \ie~$p(X, Y; \psi)$ does not share weights with $\theta$. For each example $x_i$ in a minibatch uniformly sampled from $\mathcal{D}_\text{train}$\footnote{Note that our actual formulation of $p(X, Y; \psi)$ does \textit{not} depend on $Y$, but we keep $Y$ in the notation for consistency with the formulation of the \dds~framework.}, this \dds~model outputs a scalar for the data point $x_i$. All scalars are passed through a softmax function to compute the relative probabilities of the examples in the minibatch, and their gradients are scaled accordingly when applied to $\theta$. 

We have two gradient update steps, one for the model parameter $\theta_t$ in \autoref{alg:grad_update_model} and the other for the \dds~scorer parameter $\psi$ in \autoref{alg:grad_update_p}. For the model parameter update, we can simply use any of the standard optimization update rule. For the scorer $\psi$, we use the update rule derived in \autoref{sec:diff_data_selection}.

\paragraph{Per-Example Gradient.}
In standard neural network training, a single aggregate gradient is computed with respect to a mini-batch of training data of size $n$ to improve computational efficiency. In contrast, as seen from \autoref{alg:require_per_example_grad} of \autoref{alg:image_classification_dds}, as well as from \autoref{eqn:momentum_update_for_psi}, \dds~requires us to compute $\nabla_\theta \ell(x_i, y_i; \theta_{t-1})$, the gradient for each example in a batch of training data. This potentially slows down training by a factor of $n$. A naive implementation of this operation would be very slow and memory intensive, especially when the batch size is large, \eg~our experiments on ImageNet use a batch size of $4096$ (see \autoref{sec:experiment}).  

We propose an efficient approximation of this per-example gradient computation via the first-order Taylor expansion of $\ell(x_i, y_i; \theta_{t-1})$. In particular, for any vector $v \in \mathbb{R}^{\ABS{\theta}}$, with sufficiently small $\epsilon > 0$, we have:

\begin{equation}
  \label{eqn:taylor_dot_product}
  \begin{aligned}
    & v^\top \cdot \nabla_\theta \ell(x_i, y_i; \theta_{t-1}) \\
    & \approx
    \frac{1}{\epsilon}
    \left(
      \ell\big( x_i, y_i; \theta_{t-1} + \epsilon v \big) -
      \ell\big( x_i, y_i; \theta_{t-1} \big)
    \right),
  \end{aligned}
\end{equation}
\autoref{eqn:taylor_dot_product} can be implemented by keeping a shadow version of parameters $\theta_{t-1}$, caching training loss $\ell(x_i, y_i; \theta_{t-1})$, and computing the new loss with $\theta_{t-1} + \epsilon v$. Here, $v$ is $d_\theta$ as in \autoref{alg:require_per_example_grad} of \autoref{alg:image_classification_dds}.

\subsection{\label{sec:nmt_method}Formulation for Multilingual NMT}

Next we demonstrate an application of \dds~to multilingual models for NMT, specifically for improving accuracy on low-resource languages (LRL)~\citep{nmt_transfer,rapid_adapt_nmt}.

\paragraph{Problem Setting.} In this setting, we assume that we have a particular LRL $S$ that we would like to translate into target language $T$, and we additionally have a multilingual corpus $\mathcal{D}_{\text{train}}$ that has parallel data between $n$ source languages $(S_1, S_2, ..., S_n)$ and target language $T$. We would like to pick parallel data from any of the source languages to the target language to improve translation of a particular LRL $S$, so we assume that $\mathcal{D}_{\text{dev}}$ exclusively consists of parallel data between $S$ and $T$.
Thus, \dds~will select data from $\mathcal{D}_{\text{train}}$ that improve accuracy on $S$-to-$T$ translation as represented by $\mathcal{D}_{\text{dev}}$.

\paragraph{Adaptation to NMT.} To make training more efficient and stable in this setting, we make three simplifications of the main framework in \autoref{sec:diff_data_selection} that take advantage of the problem structure of multilingual NMT.
First, instead of directly modeling $p(X,Y;\psi)$, we assume a uniform distribution over the target sentence $Y$, and only parameterize the conditional distribution of which source language sentence to pick given the target sentence: $p(X|y;\psi)$. This design follows the formulation of Target Conditioned Sampling~(TCS;~\citet{TCS}), an existing state-of-the-art data selection method that uses a similar setting but models the distribution $p(X|y)$ using heuristics.  Since the scorer only needs to model a simple distribution over training languages, we use a fully connected 2-layer perceptron network, and the input is a vector indicating which source languages are available for the given target sentence.
Second, we only update $\psi$ after updating the NMT model for a fixed number of steps.
Third, we sample the data according to $p(X|y;\psi)$ to get a Monte Carlo estimate of the objective in \autoref{eqn:psi_theta_argmin}.
This significantly reduces the training time compared to using all data. The pseudo code of the training process is in \autoref{alg:nmt_dds}. In practice, we use cosine distance instead of dot product to measure the gradient alignment between the training and dev language, because cosine distance has smaller variance and thus makes the scorer update more stable.


\begin{algorithm}
\SetAlgoLined
\DontPrintSemicolon
\SetKwInOut{Input}{Input}
\SetKwInOut{Output}{Output}
\SetCommentSty{itshape}
\SetKwComment{Comment}{$\triangleright$\ }{}

\Input{$\mathcal{D}_{\text{train}}$; K: number of data to train the NMT model before updating $\psi$; 
E: number of updates for $\psi$; 
$\alpha_1$,$\alpha_2$: discount factors for the gradient}
\Output{The converged NMT model $\theta^*$}

  Initialize $\psi_0$, $\theta_0$
  
  \Comment{Initialize the gradient of each source language}
  $grad[S_i] \leftarrow 0$ \textbf{for} \textit{i in n}
  %
  %
 
  \While{$\theta$ not converged}{
    $X, Y \leftarrow \text{load\_data}(\psi, \mathcal{D}_{\text{train}}, K)$  \label{alg:load_nmt}
  
    \Comment{Train the NMT model}
    \For{ $x_i, y$ in $X, Y$}{
      $\theta_t \leftarrow \text{GradientUpdate}\left( \theta_{t-1}, \nabla_{\theta_{t-1}} \ell(x_i, y; \theta_{t-1}) \right)$
        
      $grad[S_i] \leftarrow \alpha_1 \times \text{grad}[S_i] + \alpha_2 \times \nabla_{\theta_{t-1}} \ell(x_i, y; \theta_{t-1})$
    }
        
    
    \Comment{Optimize $\psi$}
    \For{ iter in E}{
      
      sample $B$ data pairs from $\mathcal{D}_{\text{train}}$
      
      $d_\psi \leftarrow \frac{1}{B} \sum_{j=1}^B \sum_{i=1}^n  \Big[ \text{grad}[S_i]^\top \text{grad}[S] \cdot \nabla_{\psi_{t-1}} \text{log}\left( p\left( S_i|y_j;\psi_{t-1} \right) \right) \Big]$
       
      $\psi_t \leftarrow \text{GradientUpdate}(\psi_{t-1}, d_{\psi_{t-1}})$ 
    }
  }
  \caption{\label{alg:nmt_dds}Training multilingual NMT with \dds.}
\end{algorithm}



\begin{table*}[t]
  \caption{\label{tab:results}Results for image classification accuracy (left) and multilingual MT BLEU (right). For MT, the statistical significance is indicated with $*$ (p $<$ 0.005) and $\dagger$ (p $<$ 0.0001). \dds~ outperforms the best baseline in all settings. For both image classification and NMT, \dds~performs better than other intelligent data selection methods.}
  \vspace{0.2cm}
   \begin{minipage}{.59\linewidth}
    \resizebox{\textwidth}{!}{
      \begin{tabular}{lcccc}
        \toprule
        \multirow{2}{*}{\textbf{Methods}}
        & \multicolumn{2}{c}{CIFAR-10 (WRN-28-$k$)} & \multicolumn{2}{c}{ImageNet (ResNet-50)} \\
        \cmidrule(lr){2-3} \cmidrule(lr){4-5}
        & 4K, $k=2$ & Full, $k=10$ & 10\% & Full \\
        \midrule
        Uniform &
        82.60$\pm$0.17 &
        95.55$\pm$0.15 & 
        56.36/79.45 &
        76.51/93.20 \\
        SPCL &
        81.09$\pm$0.22 &
        93.66$\pm$0.12 & 
        - &
        - \\
        BatchWeight &
        79.61$\pm$0.50 &
        94.11$\pm$0.18 & 
        - &
        - \\
        MentorNet &
        83.11$\pm$0.62 &
        94.92$\pm$0.34 & 
        - &
        - \\
        \midrule
        \dds     &
        83.63$\pm$ 0.29 &
        96.31$\pm$ 0.13 &
        \textbf{56.81}/\textbf{79.51} &
        \textbf{77.23}/\textbf{93.57} \\
         retrained \dds     &
        \textbf{85.56}$\pm$\textbf{0.20} &
        \textbf{97.91}$\pm$\textbf{0.12} &
        - &
        - \\       
        \bottomrule
      \end{tabular}
      }
      \end{minipage}
      \hfill
  \begin{minipage}{.4\linewidth}
    \resizebox{\textwidth}{!}{
      \begin{tabular}{l|cccc}
        \toprule
        \textbf{Methods} & \textbf{aze} & \textbf{bel} & \textbf{glg} & \textbf{slk} \\
        \midrule
        Uniform & 10.31 & 17.21 & 26.05 & 27.44 \\
        SPCL & 9.07 & 16.99 & 23.64 & 21.44 \\
        Related & 10.34 & 15.31 & 27.41 & 25.92 \\
        TCS     & 11.18 & 16.97 & 27.28 & 27.72 \\
        \midrule
        \dds     & 10.74 & 17.24 & 27.32 & $\mathbf{28.20^*}$ \\
        TCS+\dds & $\mathbf{11.84^*}$ & $\mathbf{17.74^\dagger}$ & \textbf{27.78} & 27.74 \\
        \bottomrule
      \end{tabular}
      }
      \end{minipage}
\end{table*}

\section{\label{sec:experiment}Experiments}
We now discuss experimental results on both image classification, an instance of the general classification problem using \autoref{alg:image_classification_dds}, and multilingual NMT using \autoref{alg:nmt_dds}\footnote{Code for image classification: \url{https://github.com/google-research/google-research/tree/master/differentiable_data_selection}. Code for NMT: \url{https://github.com/cindyxinyiwang/DataSelection}}.

\subsection{\label{exp:settings}Experimental Settings}


\noindent \textbf{Data.} We apply our method on established benchmarks for image classification and multilingual NMT.
For image classification, we use CIFAR-10~\citep{cifar10} and ImageNet~\citep{imagenet}. For each dataset, we consider two settings: a reduced setting where only roughly 10\% of the training labels are used, and a full setting, where all labels are used. Specifically, the reduced setting for CIFAR-10 uses the first $4000$ examples in the training set, and with ImageNet, the reduced setting uses the first $102$ TFRecord shards as pre-processed by~\citet{imagenet_generalize_better}. We use the size of $224 \times 224$ for ImageNet.

For multilingual NMT, we use the 58-language-to-English TED dataset~\citep{ted_pretrain_emb}. 
Following prior work~\citep{ted_pretrain_emb,rapid_adapt_nmt,sde}, we evaluate translation from four low-resource languages~(LRL) Azerbaijani~(\texttt{aze}), Belarusian~(\texttt{bel}), Galician~(\texttt{glg}), and Slovak~(\texttt{slk}) to English, where each is paired with a similar high-resource language Turkish~(\texttt{tur}), Russian~(\texttt{rus}), Portugese~(\texttt{por}), and Czech~(\texttt{ces}) (details in \autoref{app:nmt_data}).
We combine data from all 8 languages, and use \dds~to optimize data selection for each LRL.

To update the scorer, we construct $\mathcal{D}_\text{dev}$ so that it does not overlap with $\mathcal{D}_\text{test}$. For image classification, we hold out $10\%$ of the training data as $\mathcal{D}_\text{dev}$; while for multilingual NMT, we simply use the dev set of the LRL as $\mathcal{D}_\text{dev}$.

\noindent \textbf{Models and Training Details.}
For image classification, on CIFAR-10, we use the pre-activation WideResNet-28~\citep{wide_res_net}, with width factor $k=2$ for the reduced setting and $k=10$ for the normal setting. For ImageNet, we use the post-activation ResNet-50~\citep{res_net}. 
These implementations reproduce the numbers reported in the literature~\citep{wide_res_net,res_net,resnext}, and additional details can be found in \autoref{app:image_hparam}.

For NMT, we use a standard LSTM-based attentional baseline \citep{attention}, which is similar to previous models used in low-resource scenarios on this dataset~\citep{rapid_adapt_nmt,sde} and others~\citep{lownmt19} due to its relative stability compared to other options such as the Transformer \citep{vaswani2017attention}. Accuracy is measured using BLEU score \citep{bleu}.
More experiment details are noted in \autoref{app:nmt_hparam}.

\noindent \textbf{Baselines and Our Methods.}
For both image classification and multi-lingual NMT, we compare the following data selection methods. \textbf{Uniform}: data is selected uniformly from all of the data that we have available, as is standard in training models. \textbf{SPCL}~\citep{spcl}: a curriculum learning method that dynamically updates the curriculum to focus more on the ``easy'' training examples based on model loss. \textbf{\dds}: our proposed method.

For image classification, we compare with several additional methods designed for filtering noisy data on CIFAR-10, where we simply consider the dev set as the clean data. \textbf{BatchWeight}~\citep{learn_reweight}: a method that scales example training loss in a batch with a locally optimized weight vector using a small set of clean data. \textbf{MentorNet}~\citep{mentornet}: a curriculum learning method that trains a mentor network to select clean data based on features from both the data and the main model.

For machine translation, we also compare with two state-of-the-art heuristic methods for multi-lingual data selection. \textbf{Related}: data is selected uniformly from the target LRL and a linguistically related HRL \citep{rapid_adapt_nmt}. \textbf{TCS}: a recently proposed method of ``target conditioned sampling'', which uniformly chooses target sentences, then picks which source sentence to use based on heuristics such as word overlap \citep{TCS}. Note that both of these methods take advantage of structural properties of the multi-lingual NMT problem, and do not generalize to other problems such as classification. 

\paragraph{DDS with Prior Knowledge} \dds~is a flexible framework to incorporate prior knowledge about the data using the scorer network, which can be especially important when the data has certain structural properties such as language or domain. We test such a setting of \dds~for both tasks.

For image classification, we use \textbf{retrained \dds}, where we first train a model and scorer network using the standard \dds~till convergence. The trained scorer network can be considered as a good prior over the data, so we use it to train the final model from scratch again using \dds. For multilingual NMT, we experiment with \textbf{TCS+\dds}, where we initialize the parameters of \dds~with the TCS heuristic, then continue training.



\subsection{Main Results}

The results of the baselines and our method are listed in \autoref{tab:results}.
First, comparing the standard baseline strategy of ``Uniform'' and the proposed method of ``\dds'' we can see that in all 8 settings \dds~improves over the uniform baseline. This is a strong indication of both the consistency of the improvements that \dds~can provide, and the generality -- it works well in two very different settings. Next, we find that \dds~outperforms SPCL by a large margin for both of the tasks, especially for multilingual NMT. This is probably because SPCL weighs the data only by their easiness, while ignoring their relevance to the dev set, which is especially important in settings where the data in the training set can have very different properties such as the different languages in multilingual NMT. 

\dds~also brings improvements over the state-of-the-art intelligent data utilization methods. For image classification, \dds~outperforms MentorNet and BatchWeight on CIFAR-10 in all settings. For NMT, in comparison to Related and TCS, vanilla \dds~performs favorably with respect to these state-of-the-art data selection baselines, outperforming each in 3 out of the 4 settings (with exceptions of slightly underperforming Related on \texttt{glg} and TCS on \texttt{aze}).
In addition, we see that incorporating prior knowledge into the scorer network leads to further improvements. For image classification, retrained \dds~can significantly improve over regular \dds, leading to the new state-of-the-art result on the CIFAR-10 dataset. For mulitlingual NMT, TCS+\dds~achieves the best performance in three out of four cases (with the exception of \texttt{slk}, where vanilla \dds~already outperformed TCS).\footnote{Significance tests~\citep{significance_nmt} find significant gains over the baseline for \texttt{aze}, \texttt{slk}, and \texttt{bel}. For \texttt{glg}, \dds~without heuristics performs as well as the TCS baseline.}

\dds~does not incur much computational overhead. For image classification and multilingual NMT respectively, the training time is about $1.5\times$ and $2\times$ the regular training time without \dds. This contrasts favorably to previous methods that learn to select data using reinforcement learning. For example, in the IMDB movie review experiment in  \citet{learn_what_data_learn}, the data agent is trained for 200 episode, where each episode uses around 40\% of the whole dataset, requiring 80x more training time than a single training run. 
\begin{figure*}
    \includegraphics[width=\textwidth]{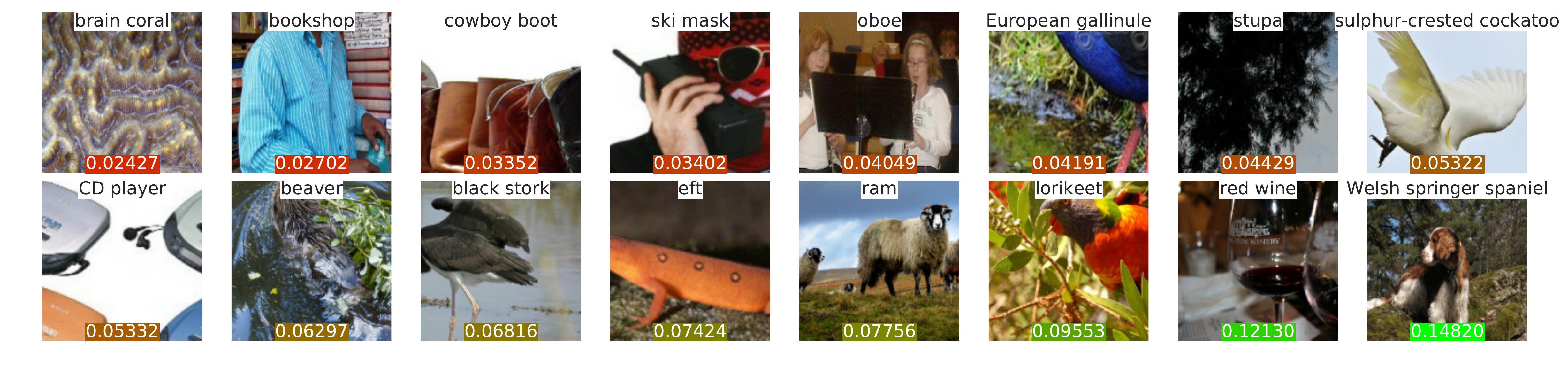}
  \caption{\label{fig:dds_score} Example images from the ImageNet and their weights assigned by \dds. A trained DDS scorer assigns higher probabilities to images from ImageNet, in which the class content is more clear. Each image's label and weight in the minibatch is shown.}
\end{figure*}

\begin{figure}
  \centering
    \includegraphics[width=0.4\textwidth]{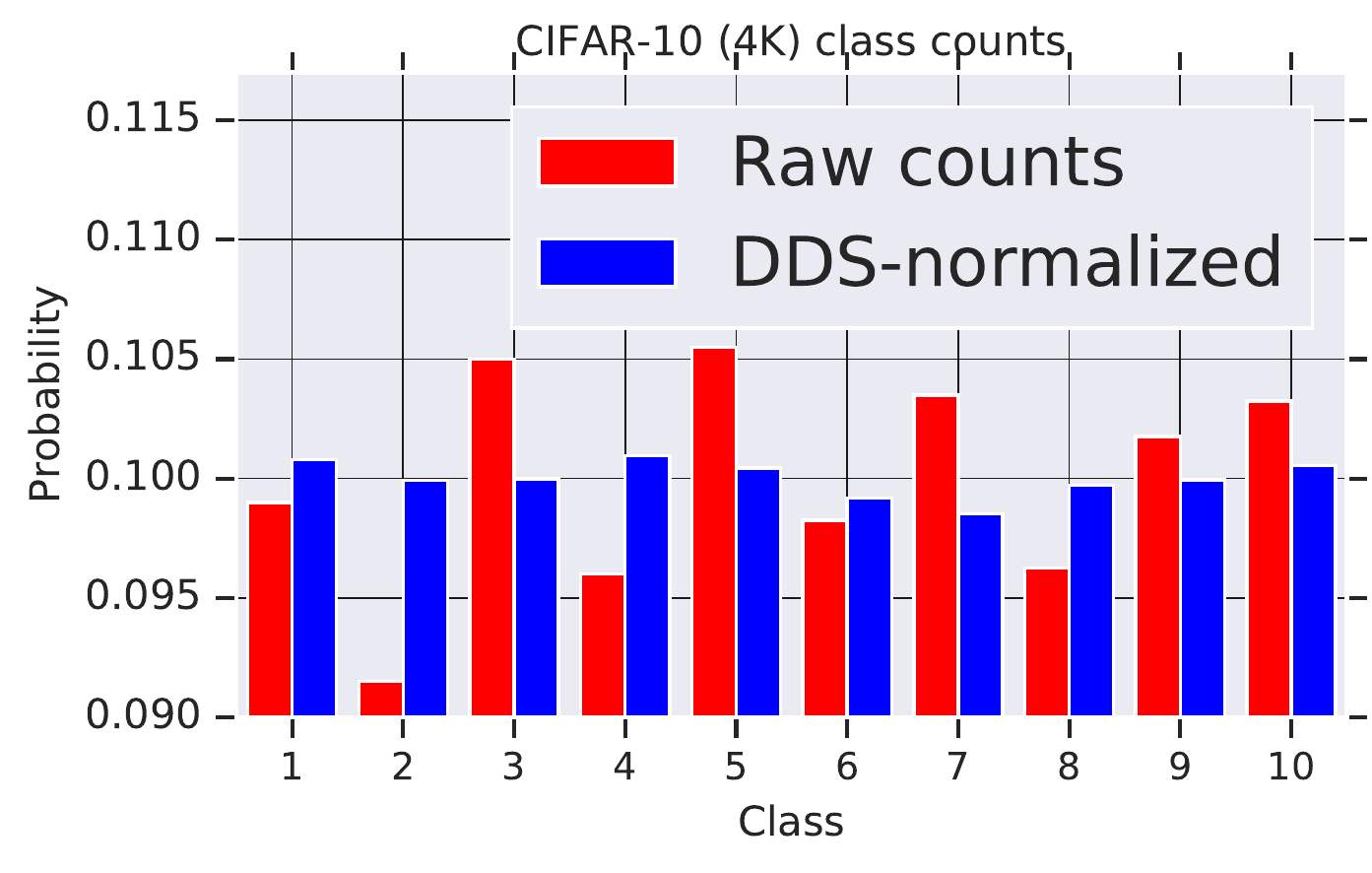}
  \caption{\label{fig:dds_distribution}A trained DDS scorer learns to balance the class distributions of CIFAR-10 4K.}
\end{figure}

\begin{figure}
    \centering
    \includegraphics[width=0.9\columnwidth]{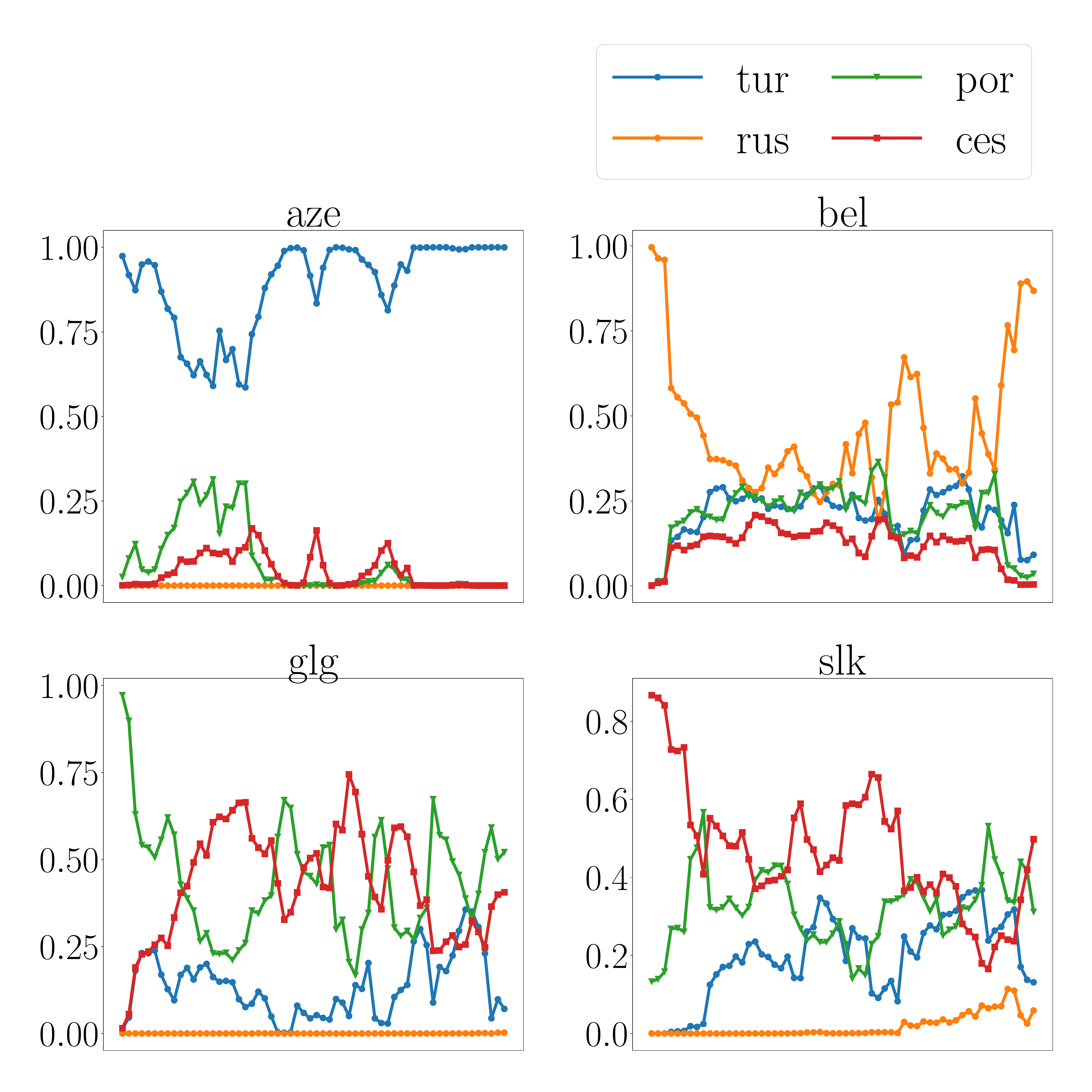}
    \caption{\label{fig:nmt_distrib_hs}Language usage for TCS$+$\dds{} by training step. The distribution is initialized to focus on the most related HRL, and \dds~learns to have a more balanced usage of all languages.}
\end{figure}

\subsection{Analysis}

\textbf{Image Classification.} Prior work on heuristic data selection has found that models perform better when fed higher quality or more domain-relevant data  towards the end of training~\citep{dynamic_data_selection_nmt,dynamic}. Here we verify this observation by analyzing the learned importance weight at the end of training for image classification. \autoref{fig:dds_distribution} shows that at the end of training, \dds~learns to balance the class distribution, which is originally unbalanced due to the dataset creation. \autoref{fig:dds_score} shows that at the end of training, \dds~assigns higher probabilities to images with clearer class content from ImageNet. These results show that \dds~learns to focus on higher quality data towards the end of training.

\textbf{NMT.}
Next, we focus on multilingual NMT, where the choice of data directly corresponds to picking a language, which has an intuitive interpretation. Since \dds~adapts the data weights dynamically to the model throughout training, here we analyze how the dynamics of learned weights.

\begin{figure}
    \centering
    \includegraphics[width=0.9\columnwidth]{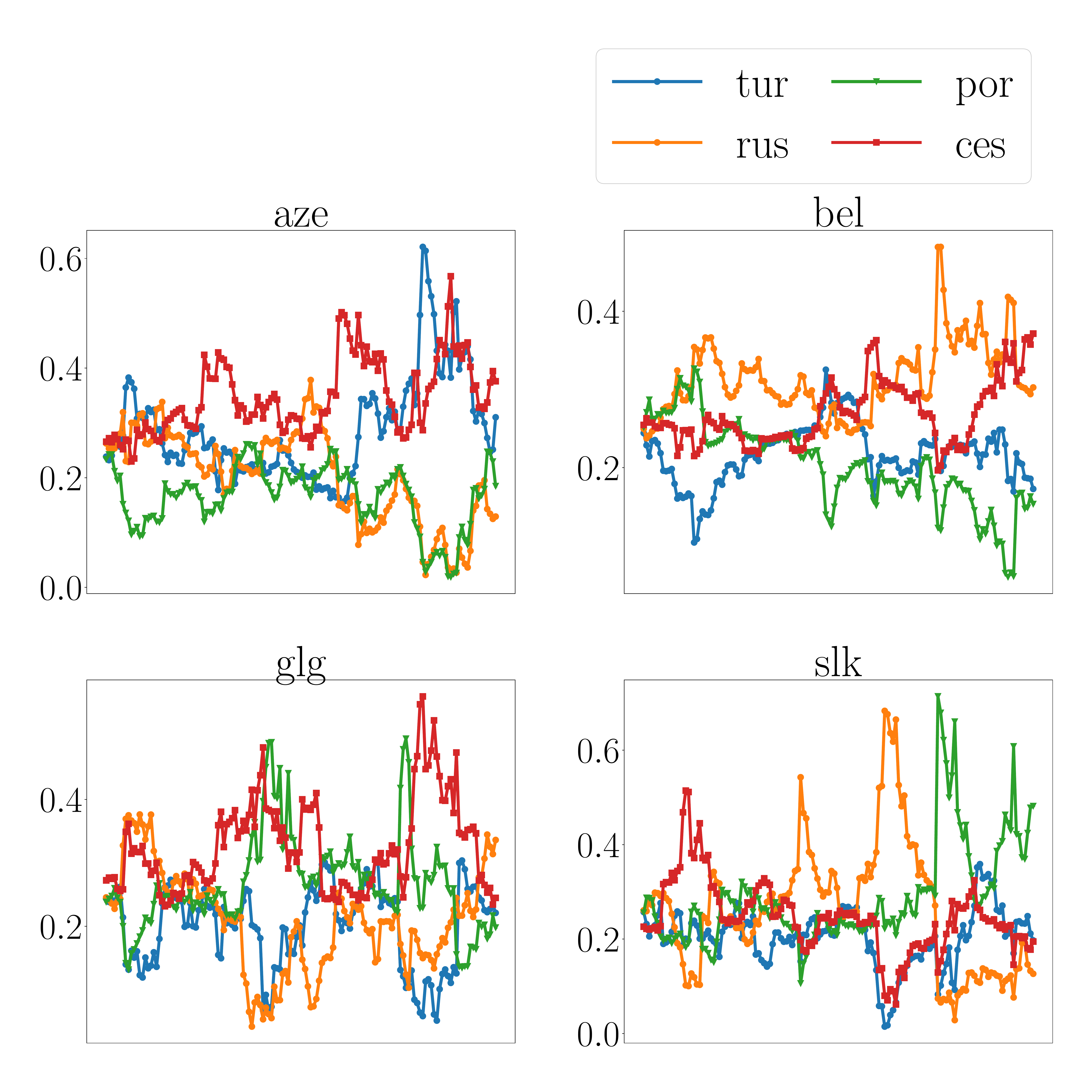}
    \caption{\label{fig:nmt_distrib_uni}Language usage for \dds~by training step. \dds~learns to upweight the most related HRL after certain training steps.}
    \vspace{-6mm}
\end{figure}
We plot the probability distribution of the four HRLs (because they have more data and thus larger impact on training) over the course of training.  \autoref{fig:nmt_distrib_hs} shows the change of language distribution for TCS+DDS. Since TCS selects the language with the largest vocabulary overlap with the LRL, the distribution is initialized to focus on the most related HRL. For all four LRLs, the percentage of their most related HRL starts to decrease as training continues. For \texttt{aze}, \dds~quickly comes back to using its most related HRL. For \texttt{gig} and \texttt{slk}, DDS learns to mainly use both \texttt{por} and \texttt{ces}, their corresponding HRL. However, for \texttt{bel}, \dds~continues the trend of using all four languages. This shows that \dds~is able to maximize the benefits of the multilingual data by having a more balanced usage of all languages.


\autoref{fig:nmt_distrib_uni} shows a more interesting trend of \dds~without heuristic initialization.
For both \texttt{aze} and \texttt{bel}, \dds~focuses on the most related HRL after a certain number of training updates.
Interestingly, for \texttt{bel}, \dds~learns to focus on both \texttt{rus}, its most related HRL, and \texttt{ces}. Similarly for \texttt{slk}, \dds~also learns to focus on \texttt{ces}, its most related HRL, and \texttt{rus}, although there is little vocabulary overlap between \texttt{slk} and \texttt{rus}.
Also notably, the ratios change  significantly over the course of training, indicating that different types of data may be more useful during different learning stages.



\section{\label{sec:related_work}Related Work}
\paragraph{Data Selection Methods} Data selection for domain adaptation for disparate tasks has also been extensively studied~\citep{moore2010intelligent,axelrod2011domain,domain_adapt_transfer,jiang-zhai-2007-instance,foster-etal-2010-discriminative,wang-etal-2017-instance}. These methods generally design heuristics to measure domain similarity,
while \dds~is a more general data selection framework that works for both classification and other usage cases. Besides domain adaptation, data selection also benefits training in the face of noisy or otherwise undesirable data~\citep{vyas-etal-2018-identifying,pham-etal-2018-fixing}. 
The idea of selecting training data that are similar to dev data has been used in works on submodular optimization~\citep{submodular_mt,learn_mix_submodular}, but they rely on features specific to the task, while \dds~directly optimizes the the dev set performance, and is generalizable across tasks. Moreover, unlike \dds, these methods cannot adaptively change the data selection scheme.

\paragraph{Instance Weighting Methods} Our method is also related to works on training instance weighting~\citep{importance_weight,learn_reweight,jiang-zhai-2007-instance,domain_adapt_transfer}. These methods reweigh data based on a manually computed weight vector, instead of using a parameterized neural network.
Notably,~\citet{learn_reweight} tackles noisy data filtering for image classification, by using meta-learning to calculate a locally optimized weight vector for each batch of data.
In contrast, our work focuses on the general problem of optimizing data usage. We train a parameterized scorer network that optimizes over the entire data space, which can be essential in preventing overfitting mentioned in ~\autoref{sec:method};  empirically our method outperform \cite{learn_reweight} by a large margin in ~\autoref{sec:experiment}. \cite{importance_weight} optimizes data weights by minimizing the error rate on the dev set. However, they use a single number to weigh each subgroup of augmented data, and their algorithm requires an expensive heuristic method to update data weights; while \dds~uses a more expressive parameterized neural network to model the individual data weights, which are efficiently updated by directly differentiating the dev loss.
 
 \paragraph{Curriculum Learning} Many machine learning approaches consider how to best present data to models. First, difficulty-based curriculum learning estimates the presentation order based on heuristic understanding of the hardness of examples~\citep{cl_bengio,SpitkovskyAJ10,baysian_curriculum,zhang2016boosting,automate_cl_GravesBMMK17,zhang2018empirical,platanios19naacl}. These methods, though effective, often generalize poorly because they require task-specific difficulty measures. On the other hand, self-paced learning~\citep{spl_kumar,spl_visual_category} defines the hardness of the data based on the loss from the model, but is still based on the assumption that the model should learn from easy examples. Our method does not make these assumptions. 

 \paragraph{RL for Training Data Usage} Our method is closest to the learning to teach framework~\citep{learn_to_teach} but their formulation involves manual feature design and requires expensive multi-pass optimization. Instead, we formulate our reward using bi-level optimization, which has been successfully applied for a variety of other tasks~\citep{bilevel_optim,hier_optim,darts,hyper_grad,learn_reweight}. \citep{reinforce_cotrain,rl_nmt,learn_active_learn} propose RL frameworks for specific natural language processing tasks, but their methods are less generalizable and requires more complicated featurization.







\section{\label{sec:conclusion}Conclusion}
We present \emph{differentiable data selection}, an efficient RL framework for optimizing training data usage. We parameterize the scorer network as a differentiable function of the data, and provide an intuitive reward function for efficiently training the scorer network. We formulate two algorithms under the \dds~framework for two realistic and very different tasks, image classification and multilingual NMT, which lead to consistent improvements over strong baselines.


\section*{Acknowledgement}
The authors would like to thank Quoc Le for various suggestions, and thank Hanxiao Liu for comments on the paper's first draft. The first author Xinyi Wang is supported by a research grant
from the Tang Family Foundation. The
authors would like to thank Amazon for providing
GPU credits.

\bibliography{main}
\bibliographystyle{icml2019}

\newpage
\appendix
\section{\label{app} Appendix}

\subsection{\label{app:grad_of_optimizers}Deriving gradient of $\psi$ for Different Optimizers}
First, we rewrite the update rule of $\theta$ in \autoref{eqn:theta_update} to incorporate the effect of its specific optimization algorithm.

For a fixed value of $\psi$, $J(\theta, \psi)$ can be optimized using a stochastic gradient update. Specifically, at time step $t$, we update
\begin{equation}
  \label{eqn:theta_update_rule}
   \small
  \begin{aligned}
    \theta_t \leftarrow \theta_{t-1} - g\big( \nabla_\theta J(\theta_{t-1}, \psi) \big)
  \end{aligned}
\end{equation}
where $g(\cdot)$ is any function that may be applied to the gradient $\nabla_\theta J(\theta_{t-1}, \psi)$. For instance, in standard gradient descent $g(\cdot)$ is simply a linear scaling of $\nabla_\theta J(\theta_{t-1}, \psi)$ by a learning rate $\eta_t$, while with the Adam optimizer~\citep{adam} $g$ also modifies the learning rate on a parameter-by-parameter basis.

Due to the relationship between $\theta_t$ and $\psi$ as in \autoref{eqn:theta_update_rule}, $J(\theta_t, \mathcal{D}_\text{dev})$ is differentiable with respect to $\psi$. 
By the chain rule, we can compute the gradient $\nabla_\psi J(\theta_t, \mathcal{D}_\text{dev})$ as follows:
\begin{equation}
  \label{eqn:two_step_update_general}
   \small
  \begin{aligned}
      &\text{(chain rule):} \\
    \nabla_\psi J(\theta_t, \mathcal{D}_\text{dev})
      &= \nabla_{\theta_t} J(\theta_t, \mathcal{D}_\text{dev})^\top \cdot \nabla_\psi \theta_t(\psi) \\
      &\text{(substitute $\theta_t$ from \autoref{eqn:theta_update_rule}):} \\
      &= \nabla_{\theta_t} J(\theta_t, \mathcal{D}_\text{dev})^\top \cdot \nabla_\psi \left( \theta_{t-1} - g\big( \nabla_\theta J(\theta_{t-1}) \big) \right) \\
      &\text{(assume $\nabla_\psi \theta_{t-1} \approx 0$)} \\
      &\approx -\nabla_{\theta_t} J(\theta_t, \mathcal{D}_\text{dev})^\top \cdot \nabla_\psi g\big( \nabla_\theta J(\theta_{t-1}) \big)  
  \end{aligned}
\end{equation}
Here, we make a Markov assumption that $\nabla_\psi \theta_{t-1} \approx 0$, assuming that at step $t$, given $\theta_{t-1}$ we do not care about how the values of $\psi$ from previous steps led to $\theta_{t-1}$. \autoref{eqn:two_step_update_general} leads to a rule to update $\psi$ using gradient descent:
\begin{equation}
  \label{eqn:psi_update_rule}
   \small
  \begin{aligned}
    \psi_{t+1} 
      &\leftarrow \psi_t + \eta_\psi \nabla_{\theta_t} J(\theta_t, \mathcal{D}_\text{dev})^\top \cdot \nabla_\psi g\big( \nabla_\theta J(\theta_{t-1}, \psi_t) \big),
  \end{aligned}
\end{equation}

Here we first derive $\nabla_\psi g$ for the general stochastic gradient descent~(SGD) update, then provide examples for two other common optimization algorithms, namely Momentum~\citep{nesterov} and Adam~\citep{adam}.

\paragraph{SGD Updates.} The SGD update rule for $\theta$ is as follows
\begin{equation}
  \label{eqn:sgd_update}
   \small
  \begin{aligned}
    \theta_t &\leftarrow \theta_{t-1} - \eta_t \nabla_\theta J(\theta_{t-1}, \psi)
  \end{aligned}
\end{equation}
where $\eta_t$ is the learning rate. Matching the updates in \autoref{eqn:sgd_update} with the generic framework in \autoref{eqn:theta_update_rule}, we can see that $g$ in \autoref{eqn:theta_update_rule} has the form: 
\begin{equation}
  \label{eqn:momentum_update_g}
   \small
  \begin{aligned}
    g\big(\nabla_\theta J(\theta_{t-1}, \psi)\big) = \eta_t \nabla_\theta J(\theta_{t-1}, \psi)
  \end{aligned}
\end{equation}
This reveals a linear dependency of $g$ on $\nabla_\theta J(\theta_{t-1, \psi})$, allowing the exact differentiation of $g$ with respect to $\psi$. From \autoref{eqn:psi_update_rule}, we have
\begin{equation}
  \label{eqn:momentum_update_for_psi}
   \small
  \begin{aligned}
    &\nabla J(\theta_t, \mathcal{D}_\text{dev})^\top \cdot \nabla_\psi g\big( \nabla_\theta J(\theta_{t-1}, \psi) \big) \\
    &= \eta_t \cdot \nabla_\psi \mathbb{E}_{x, y \sim p(X, Y; \psi)} \left[\nabla J(\theta_t, \mathcal{D}_\text{dev})^\top \cdot \nabla_\theta \ell(x, y; \theta_{t-1} )\right] \\
    &= \eta_t \mathbb{E}_{x, y \sim p(X, Y; \psi)} \left[\left(\nabla J(\theta_t, \mathcal{D}_\text{dev})^\top \cdot \nabla_\theta \ell(x, y; \theta_{t-1} ) \right) \cdot \nabla_\psi \log{p(x, y; \psi)} \right]
  \end{aligned}
\end{equation}
Here, the last equation follows from the log-derivative trick in the REINFORCE algorithm~\citep{reinforce}. We can consider the alignment of dev set and training data gradients as the reward for update $\psi$. In practice, we found that using cosine distance is more stable than simply taking dot product between the gradients. Thus in our implementation of the machine translation algorithm, we use $\text{cos}\left(J(\theta_t, \mathcal{D}_\text{dev})^\top \cdot \nabla_\theta \ell(x, y; \theta_{t-1} ) \right)$ as the reward signal.

\paragraph{Momentum Updates.} The momentum update rule for $\theta$ is as follows
\begin{equation}
  \label{eqn:momentum_update}
   \small
  \begin{aligned}
    m_t &\leftarrow \mu_t m_{t-1} + \eta_t \nabla_\theta J(\theta_{t-1}, \psi) \\
    \theta_t &\leftarrow \theta_{t-1} - m_t,
  \end{aligned}
\end{equation}
where $\mu_t$ is the momentum coefficient and $\eta_t$ is the learning rate. This means that $g$ has the form:
\begin{equation}
  \label{eqn:momentum_update_g}
   \small
  \begin{aligned}
    g(x) &= \mu m_{t-1} + \eta_t x \\
    g'(x) &= \eta_t
  \end{aligned}
\end{equation}
Therefore, the computation of the gradient $\nabla_{\psi}$ for the Momentum update is exactly the same with the standard SGD update rule in \autoref{eqn:momentum_update_for_psi}.

\paragraph{Adam Updates.} We use a slightly modified update rule based on Adam~\citep{adam}:
\begin{equation}
  \label{eqn:adam_update}
   \small
  \begin{aligned}
    &g_t \leftarrow \nabla_\theta J(\theta_{t-1}, \psi) \\
    &v_t \leftarrow \beta_2 v_{t-1} + (1 - \beta_2) g_t^2 \\
    &\hat{v}_t \leftarrow v_t / (1 - \beta_2^t) \\
    &\theta_t \leftarrow \theta_{t-1} - \eta_t \cdot g_t / \sqrt{\hat{v}_t + \epsilon}
  \end{aligned}
\end{equation}
where $\beta_2$ and $\eta_t$ are hyper-parameters. This means that $g$ is a component-wise operation of the form:
\begin{equation}
  \label{eqn:adam_update_g}
   \small
  \begin{aligned}
    g(x) &= \frac{\eta_t \sqrt{1 - \beta_2^t} \cdot x}{\sqrt{\beta_2 v_{t-1} + (1 - \beta_2) x^2 + \epsilon}} \\
    g'(x) &= \frac{\eta_t \sqrt{1 - \beta_2^t} (\beta_2 v_{t-1} + \epsilon)}{\big( \beta_2 v_{t-1} + (1 - \beta_2) x^2 + \epsilon \big)^{3/2}} \approx \eta_t \sqrt{\frac{1 - \beta_2^t}{\beta_2 v_{t-1}}},  
  \end{aligned}
\end{equation}
the last equation holds because we assume $v_{t-1}$ is independent of $\psi$. Here the approximation makes sense because we empirically observe that the individual values of the gradient vector $\nabla_\theta J(\theta_{t-1}, \psi)$,~\ie~$g_t$, are close to $0$. Furthermore, for Adam, we usually use $\beta_2 = 0.999$. Thus, the value $(1 - \beta_2) x^2$ in the denominator of \autoref{eqn:adam_update_g} is negligible. With this approximation, the computation of the gradient $\nabla_\psi$ is almost the same with that for SGD in \autoref{eqn:momentum_update_for_psi}, with one extra component-wise scaling by the term in \autoref{eqn:adam_update_g}.

\subsection{\label{app:nmt_hparam} Hyperparameters for multilingual NMT}
In this section, we give a detailed description of the hyperparameters used for the multilingual NMT experiments.
\begin{itemize}
    \item We use a 1 layer LSTM with hidden size of 512 for both the encoder and decoder, and set the word embedding to size 128.
    \item For multilingual NMT, we only use the scorer to model the distribution over languages. Therefore, we use a simple 2-layer perceptron network as the scorer architecture. Suppose the training data is from $n$ different languages. For each target sentence and its corresponding source sentences, the input feature is a $n$-dimensional vector of 0 and 1, where 1 indicates a source language exists for the given target sentence.
    \item We simply use the dev set that comes with the dataset as $\mathcal{D}_\text{dev}$ to update the scorer.
    \item The dropout rate is set to 0.3.
    \item For the NMT model, we use Adam optimizer with learning rate of 0.001. For the distribution parameter $\psi$, we use Adam optimizer with learning rate of 0.0001.
    \item We train all models for 20 epochs without any learning rate decay.
    \item We optimize both the NMT and \dds~models with Adam, using learning rates of 0.001 and 0.0001 for $\theta$ and $\psi$ respectively.
\end{itemize}

\subsection{\label{app:nmt_data} Dataset statistics for Multilingual NMT}
\begin{table}[H]
  \centering
  \begin{tabular}{c|ccc|cc}
  \toprule
  \textbf{LRL} & \textbf{Train} & \textbf{Dev} & \textbf{Test} & \textbf{HRL} & \textbf{Train} \\
  \midrule
  aze & 5.94k &  671 &  903 & tur & 182k \\
  bel & 4.51k &  248 &  664 & rus & 208k \\
  glg & 10.0k &  682 & 1007 & por & 185k \\
  slk & 61.5k & 2271 & 2445 & ces & 103k \\
  \bottomrule
  \end{tabular}
  \vspace{0.2cm}
  \caption{\label{tab:nmt_data}Statistics of the multilingual NMT datasets.}
\end{table} 

\subsection{\label{app:image_hparam} Hyperparameters for image classification}
In this section, we provide some additional details for the image classification task:
\begin{itemize}
  \item We use the cosine learning rate decay schedule~\citep{cosine_lr}, starting at $0.1$ for CIFAR-10 and $3.2$ for ImageNet, both with $2000$ warmup steps. 
  \item For image classification, we use an identical network architecture with the main model, but with independent weights and a regressor to predict the score instead of a classifier to predict image classes.
  \item To construct the $\mathcal{D}_\text{dev}$ to update the scorer, we hold out about 10\% of the \textit{training} data. For example, in CIFAR-10 (4,000), $\mathcal{D}_\text{dev}$ is the last 400 images, while in ImageNet-10\%, since we use the first 102 TFRecord shards, $\mathcal{D}_\text{dev}$ consists of the last 10 shards. Here, “last” follows the order in which the data is posted on their website for CIFAR-10, and the order in which the TFRecord shards are processed for ImageNet. All data in $\mathcal{D}_\text{dev}$ are excluded from $\mathcal{D}_\text{train}$. Thus, for example, with CIFAR-10 (4,000), $|\mathcal{D}_\text{train}| = 3600$, ensuring that in total, we are only using the amount of data that we claim to use.

  \item We maintain a moving average of all model parameters with the rate of $0.999$. Following~\citet{imagenet_generalize_better}, we treat the moving statistics of batch normalization~\citep{batch_norm} as \textit{untrained parameters} and also add them to the moving averages. 
  \item  For ImageNet, we use the post-activation ResNet-50~\citep{res_net}. 
The batch sizes for CIFAR-10 and for ImageNet are $128$ and $4096$, running for 200K steps and 40K steps, respectively. 
\end{itemize}

\end{document}